\documentclass[lettersize,journal]{IEEEtran}
\usepackage{amsmath,amsfonts}
\usepackage{algorithmic}
\usepackage{algorithm}
\usepackage{array}
\usepackage{textcomp}
\usepackage{stfloats}
\usepackage{url}
\usepackage{verbatim}
\usepackage{graphicx}
\usepackage{cite}
\usepackage{algorithm}
\usepackage{algorithmic}
\usepackage{multirow}
\usepackage{booktabs}
\usepackage{amssymb}
\usepackage{subfigure}
\usepackage{flushend}
\usepackage{color}
\begin{document}

\title{Towards Unbiased Multi-label Zero-Shot Learning \\
with Pyramid and Semantic Attention}


\author{Ziming~Liu,
       Song~Guo,~\IEEEmembership{Fellow,~IEEE}, 
       Jingcai~Guo,~\IEEEmembership{Member,~IEEE,}
       Yuanyuan~Xu, 
	and~Fushuo~Huo

\thanks{Z. Liu, S. Guo and J. Guo are with the Department of Computing, The Hong Kong Polytechnic University, Hong Kong SAR., China (e-mail: ziming.liu@connect.polyu.hk, song.guo@polyu.edu.hk, jingcai.guo@connect.polyu.hk).}


\thanks{Y. Xu is with the School of Information and Software Engineering, University of Electronic Science and Technology of China, Chengdu 610054, China (e-mail: hyries@std.uestc.edu.cn).}

\thanks{F. Huo is with the State Key Laboratory of Power Transmission Equipment and System Security, Chongqing University, Chongqing 400044, China (e-mail: 20191102013t@cqu.edu.cn).}

}




\maketitle

\begin{abstract}
Multi-label zero-shot learning extends conventional single-label zero-shot learning to a more realistic scenario that aims at recognizing multiple unseen labels of classes for each input sample. 
Existing works usually exploit attention mechanism to generate the correlation among different labels. However, most of them are usually biased on several \textit{major classes} while neglect 
most of the \textit{minor classes} with the same importance in input samples, and may thus result in overly diffused attention maps that cannot sufficiently cover \textit{minor classes}. 
We argue that disregarding the connection between major and minor classes, i.e., correspond to the global and local information, respectively, is the cause of the problem. 
In this paper, we propose a novel framework of unbiased multi-label zero-shot learning, by considering various class-specific regions to calibrate the training process of the classifier. Specifically, \textit{Pyramid Feature Attention} (\textit{PFA}) is proposed to build the correlation between global and local information of samples to balance the presence of each class. 
Meanwhile, for the generated semantic representations of input samples, we propose \textit{Semantic Attention} (\textit{SA}) to strengthen the element-wise correlation among these vectors, which can encourage the coordinated representation of them. 
Extensive experiments on the large-scale multi-label zero-shot benchmarks \textit{NUS-WIDE} and \textit{Open-Images} demonstrate that the proposed method surpasses other representative methods by significant margins.


\end{abstract}


\section{Introduction}

\begin{figure}[htbp]
    \centering
    \includegraphics[width=0.42\textwidth]{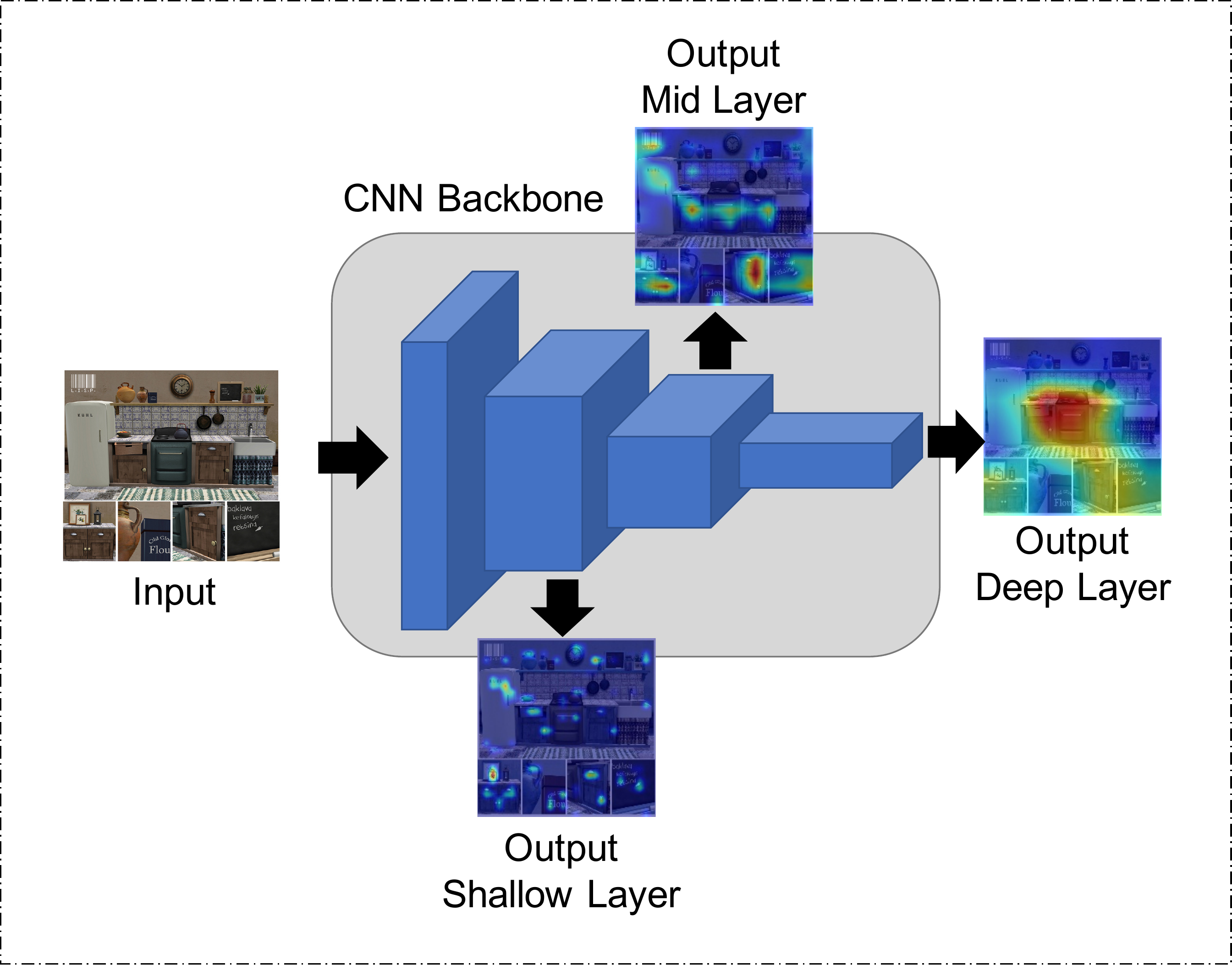}
    \caption{An illustration of the size of the object of interest (OoI) based on CNNs at different scales. Our model integrates multi-scale features to balance the correlation between minor and major classes. It is observed that minor and major classes are attended from shallow to deeper layers, gradually. In the shallow layer, the CNNs pays more attention to minor classes, such as photo frames and vases, while in the mid layer, the CNNs focuses on major classes such as door, and in the deep layer, the CNNs keeps its eye on multiple large objects and start paying attention to the whole image.}
    \label{fig1}
\end{figure}
\IEEEPARstart{T}{o} accurately recognize and describe the rich information contained within a real-world image has been a hot research topic in recent years. 
With the popularity of smart computing and machine intelligence, various devices such as mobile phones are usually equipped with photo capture and storage functions, that can produce tons of images and continue spawning the demand for recognition. 
Concretely, the multi-label classification, i.e., to recognize all labels/classes of objects in an image, is one of the fundamental and essential tasks in image recognition, which is of crucial importance to achieve several down-stream tasks such as scene analysis, autonomous driving, intelligent monitoring, and so on.
%

In the past few years, with the increasing development of deep learning techniques and tons of well-annotated training images, recent multi-label classification models \cite{wang2016cnn, gong2013deep, yu2014large, weston2011wsabie, durand2019learning,chen2019multi, feng2019collaboration} can easily achieve superior performance in both recognition accuracy and generalization ability in a close-set scenario, wherein, both training and testing images fall into the same classes of labels. 
Notably, existing multi-label classification models usually improve their recognition performance by considering the dependencies among labels, and jointly ensemble separate classifiers of each individual label for better classification. 
%
However, in real-world applications such as Flickr\footnote{https://www.flickr.com} and other online apps, billions of users are constantly releasing images containing various seen classes and many more novel unseen classes. For example, an image on Flickr can only contain incomplete tags from users, i.e., keywords that facilitate the description and searching process, which may result in less classes been recognized and degrades the recognition performance dramatically. 
In a more realistic setting, the model is necessarily to be endowed with certain predictive ability and thus to infer whether a new class exists in an image. However, conventional multi-label classification models fail to handle the task that requires not only recognizing objects that come from seen classes, but also generalizing to objects whose labels have never been seen or observed by the classifier. 
%
%
Despite that one can always label newly generated data and re-train the models to address the problem, however, such trivial and repeated labor works are very expensive and time-consuming. 

In contrast, zero-shot learning (ZSL) designs a model to simulate the way humans predict, which differs from traditional machine learning methods. 
Specifically, a shared semantic feature space between seen and novel unseen classes, e.g., word vector space~\cite{frome2013devise, socher2013zero, akata2015label, xian2016latent, norouzi2014zero}, is exploited as the bridge to fill the gap between them. 
In practice, the ZSL setting can be regarded as an extreme case of transfer learning, that can generalize well to novel concepts that haven't been observed during training. 
The taxonomy of ZSL mainly includes conventional zero-shot learning (CZSL) and generalized zero-shot learning (GZSL). 
In CZSL, the testing images are limited to unseen classes. However, this setting is not that practical in real-world application since one may also frequently encounter samples from seen classes during inference. 
In contrast, the GZSL setting requires the model to generalize well to both seen and unseen classes, which is a more challenging scenario in ZSL due to the following two reasons. 
First, the training samples only contain seen classes while the testing samples contain both seen and unseen ones, it can greatly increase the difficulty of prediction. 
Second and more importantly, if high overlap exists in the feature space between seen and unseen classes, the model may easily shift to seen classes, i.e., wrongly classify samples of unseen classes to seen classes, and degrade the recognition performance dramatically. 
%

Existing state-of-the-art ZSL models usually focus on predicting single label for each image \cite{akata2016multi, deutsch2017zero, frome2013devise, morgado2017semantically, kodirov2017semantic, li2017zero, xian2017zero, zhang2017learning, guo2020novel}, which obviously cannot present all useful information within it, and thus may not be that practical for real-world applications where the multi-label information is urgently needed. 
The challenge of multi-label prediction in ZSL mainly lies in the complex contents and relationships in a single image across various classes. 
To overcome this challenge and construct a more applicable recognition system, our interest is to investigate the multi-label zero-shot learning (ML-ZSL), that can recognize multiple objects simultaneously from seen and unseen classes in just one image. 
%
%
Recently, a few works have addressed the problem of ML-ZSL by using techniques such as attention-sharing~\cite{huynh2020shared}, structured knowledge graphs~\cite{lee2018multi}, convex combinations~\cite{norouzi2014zero}, GAN-based features synthesis~\cite{gupta2021generative}, and global feature generation~\cite{mensink2014costa, zhang2016fast}, and obtained some promising results. 
Specifically, 
Norouzi~\text{et al.}~\cite{norouzi2014zero} designed a new multi-label classifier and fused multi-label features to generate combined features to represent images. The combined image features are then fed into the word vector space to rank the labels. 
Zhang~\text{et al.}~\cite{zhang2016fast} trained the network to make the input image features have a main direction, wherein, the relevant labels are ranked better than the irrelevant labels, so as to determine the unseen classes present in the image. 
It is noticed that, the idea of the above two methods, i.e., \cite{norouzi2014zero,zhang2016fast}, is to fuse multi-class label features. However, if more classes exist in the images, the recognition accuracy may be greatly reduced due to the blurring of single-class features. 
Alternatively, \textit{LESA}~\cite{huynh2020shared} used multiple spatial attentions to do the feature extraction and sharing, which can better attend to relevant images regions. 
Differently, Gupta~\text{et al.}~\cite{gupta2021generative} introduced the generative model into the multi-label (generalized) zero-shot learning problem, which can generate multi-label visual features with consistent semantics by synthesizing attributes of different classes, to achieve the purpose of prediction. 
The above two methods, i.e., \cite{huynh2020shared,gupta2021generative}, can distinguish and fuse multiple features through the attention mechanism and the generative model, respectively, which greatly improves the model's ability. 
However, they still ignore constructing effective semantic correlation between features, which may generate less accurate semantic representations for input samples. 
Meanwhile, none of them have considered improving the recognition accuracy of seen classes. 
%
Worse still, most existing methods usually tend to overly concentrate on one or a limited number of labels, which cannot deal with all attended objects in an equal manner. 
Such a limitation may result in a biased model that can easily recognize a few major classes while neglecting most of the minor classes. Meanwhile, none of them can extensively nor try to make utilization of the semantic diversity, which may suppress the representation ability of generated semantic vectors of input samples. 

To deal with the above problems and construct an unbiased ML-ZSL model, we propose to fully make use of various class-specific regions to calibrate the model training process. Specifically, we propose \textit{Pyramid Feature Attention (PFA)} module to integrate multi-scale features into the extraction process to balance the correlation between minor and major classes (Fig.~\ref{fig1}), hence the network can pay equal attention to both of them and improves the recognition accuracy. 
Notably, despite that pyramidally augmenting the feature maps can facilitate the extraction of semantic information, while the establishment of the pyramid can substantially increase the model parameters. To balance the multi-scale extraction and computation overhead, we propose to unify the size of feature maps within our pyramid architecture by reshaping larger-scale feature maps into the same size as small ones. Such unified feature maps can significantly reduce the system complexity while retain the extraction ability for attention purpose. 
It is proved by the proposed \textit{PFA} that, the global and local information corresponds mainly to the major and minor classes, respectively. 
%
%
Moreover, we also propose \textit{Semantic Attention (SA)} module to achieve the attention sharing by considering element-wise coordination. Different from \textit{LESA}~\cite{huynh2020shared} and Gupta~\text{et al.}~\cite{gupta2021generative}, we propose to generate new semantic information by linking different classes of semantic information using the attention mechanism. 
It is proved that by jointly using the two attentions, an unbiased multi-label zero-shot recognition model can be obtained. 

In summary, our contributions are four-fold:
\begin{enumerate}
\item First, our method elaborates on the equal presence of major and minor classes, which is considered as the major contribution for existing ML-ZSL research. Concretely, previous methods are usually biased on major classes (i.e., corresponds to larger objects within an image) and may thus degrade the recognition of minor classes. To solve such an issue, we consider the multi-scale extraction in a model, where we found the shallow layers may contain more details on minor classes while the deeper layers may eventually converge to major classes due to the local reception fields and continuous poolings. As such, we apply the pyramid structures to fully make use of such multi-scale information, which is the main motivation of \textit{PFA}. Such a consideration is also the first attempt in ML-ZSL. 

\item Second, our method further elaborates on the element-wise correlation of generated semantic vectors. Previous methods usually neglect such a fine-grained and cross-class consideration on semantic vectors, and may thus generate (or partially) uncorrelated representations of samples. To solve such an issue, we consider building a semantic matrix covering the element-wise and cross-class correlations among semantic vectors in an end-to-end manner. Such a consideration can encourage the model to generate more coordinated representations of samples, which is the main motivation of \textit{SA}. 

\item Third, the proposed \textit{PFA} and \textit{SA} can be regarded as an integrated solution, where \textit{PFA} pays attention to the balance between classes to reduce bias, and \textit{SA} generates better semantic information for each class. 

\item Extensive experiments on large-scale ML-ZSL benchmark datasets, i.e., \textit{NUS-WIDE} and \textit{Open-Images}, verified the effectiveness of our method along with state-of-the-art results.
\end{enumerate}

\section{Related Work}

\subsection{Multi-Label Classification}

\noindent With the continuous development of deep learning techniques, the multi-label classification has made remarkable achievements. 
Concretely, existing methods can be roughly divided into three directions including convolutional neural network (CNN) based models such as \cite{wang2016cnn, gong2013deep, yu2014large, weston2011wsabie, durand2019learning, feng2019collaboration}, graph neural network (GNN) based models such as \cite{chen2019multi, kipf2016semi}, and recurrent neural network (RNN) based models such as \cite{yazici2020orderless, nam2017maximizing}.

The GNN-based multi-label classification aims at establishing associations between different labels through a graph network based on prior information, which is helpful to infer to novel or rarely observed objects. However, the graph propagation is significantly time-consuming and may result in overly smooth feature representations. 
The RNN-based methods can establish connections between different labels through the time series, and sorting the labels by the frequency. It can help build the relationship much faster. However, the addition of this prior information often ignores the natural attributes of labels, which eventually leads to biased prediction results.
The main-stream of CNN-based methods usually resort to explore the attention distribution of the corresponding image regions of different classes. Such kind of spatial attention mechanism can roughly find the positions corresponding to the classes, and achieve relatively good recognition performance. However, the pure spatial attention lacks the ability to generalize to unseen labels, thus it is difficult to be used to predict unseen classes. Other CNN-based methods, such as multi-label object detection~\cite{zhang2018single, ren2015faster, he2017mask, law2018cornernet}, resort to locate objects through region proposal. However, the premise is that the dataset needs to provide a bounding box containing the object as training input to locate the label, which is very time-consuming. 

\subsection{Zero-Shot Learning}
\noindent The emergence of ZSL~\cite{wang2016cnn, gong2013deep, yu2014large, weston2011wsabie, durand2019learning,chen2019multi, feng2019collaboration} is to enable the network to predict unseen classes and improve the generalization ability of the network. 
This is typically achieved by constructing and exploring the correlation between seen and unseen classes, i.e., the visual-label semantic space generated by the word vectors or the attribute vectors. 
Practically, an image containing one or more unseen classes is first fed into a classifier trained by seen classes, to obtain its semantic vector. Such a semantic vector is then sent into the semantic space to classify the unseen classes through similarity calculation with both seen and unseen classes. Recently, several advanced methods also tried to model the parts of the image using attention mechanism and obtained superior results~\cite{gao2020multi, wang2017multi, ye2020attention, you2020cross}. 
Based on the predicted labels/classes, existing ZSL can be divided into single-label zero-shot learning and multi-label zero-shot learning (ML-ZSL). 

\subsubsection{Single-label Zero-Shot Learning}
Single-label prediction is the very first and most widely studied scenario in ZSL. Early works of ZSL usually explore efficient and robust mapping between visual and semantic feature space to achieve the recognition~\cite{farhadi2009describing,frome2013devise,lampert2014attribute,shigeto2015ridge,wang2015zero,bucher2016improving,zhang2016zero,guo2019adaptive,changpinyo2016synthesized,kodirov2017semantic,guo2019ams,zhu2018generative}. Most recently, Rahman~\textit{et al.}~\cite{ZSFSL2018} developed an effective and unified method for both zero-shot and few-shot learning by inducing the concept of class adapting principal directions that make the embeddings of unseen class images in the semantic space more discriminative. 
Guo~\textit{et al.}~\cite{GUOTMM} proposed a novel model to align the manifold structures between the visual and semantic feature spaces via expansion of semantic features. 
Differently, to enhance the discriminability on both seen and unseen domains, Zhang~\textit{et al.}~\cite{ZSLTMM4} developed a systematical solution via separately learning visual prototypes and proposed an efficient solution. 
To mitigate biases towards seen classes and accommodate diverse tasks, Li~\textit{et al.}~\cite{ZSLTMM2} proposed an attribute-aware modulation network to remedy the defect of meta generative approaches, which is devoted to explore the common model shared across task distributions. 
Similarily, Min~\textit{et al.}~\cite{ZSLTMM3} proposed dual-cycle consistency and domain division constraints to make obtain the domain similarity and specialities to overcome biases. Although good results have been achieved, these methods generally lack generalization capabilities and cannot directly migrate the methods to multi-label scenarios.

\subsubsection{Multi-label Zero-Shot Learning}
In recent years, the multi-label prediction has gained increasing attention in ZSL. For example, 
Norouzi~\textit{et al.}~\cite{norouzi2014zero} proposed to directly combine the image classifier and the semantic word embedding model into a new model, and the output of the classifier is convexly combined and then sent to the semantic embedding space for prediction. 
Zhang~\textit{et al.}~\cite{zhang2016fast} used a linear mapping and a non-linear deep neural network to approximate the principal direction of images. The principal direction is regarded as the main information contained in the image. The word vector of the relevant labels of the image is ranked ahead of irrelevant labels. 
Differently, Lee~\textit{et al.}~\cite{lee2018multi} designed a new framework based on the knowledge graph to construct a knowledge graph between multiple labels. At the same time, a new information propagation mechanism is learned from semantic information to model the relationship between seen and unseen labels. 
Ji~\textit{et al.}~\cite{MZSLTIP} proposed a flexible framework, consisting of two modules: a visual-semantic regression unit and multi-label zero-shot prediction unit, which aims to embed the visual features to semantic space and solve the prediction as the ranking problem. 
Most recently, \textit{LESA}~\cite{huynh2020shared} developed a new shared attention model, which lets the unseen labels select among a set of shared attentions. Although this model can focus the network on key areas, it lacks the integration of global and local information, so the class balance is affected. In addition, it has a very complicated loss function, which is not conducive to improvements. 
Gupta~\textit{et al.}~\cite{gupta2021generative} proposed a generative method to explore the multi-label feature synthesis problem via integrating three different fusion solutions with two representative generative architectures, to compute reliable attention maps for unseen classes.

\subsubsection{Our Method}
In this paper, we observe that existing methods usually fail or degrade their performance in predicting images that with more minor classes. We argue the global and local information, i.e., corresponds to major and minor classes, should be given the same importance to balance and calibrate the recognition on minor classes and further improve the overall results towards unseen classes. 
To the best of our knowledge, we are the first to jointly consider the multi-scales of feature extraction and element-wise coordination of generated semantic vectors, which can well mitigate the minor class recognition degradation and increase the semantic diversity. Moreover, our method can achieve state-of-the-art results without a large-scale backbone network.


\begin{figure*}
    \centering
    \includegraphics[width=0.8\textwidth]{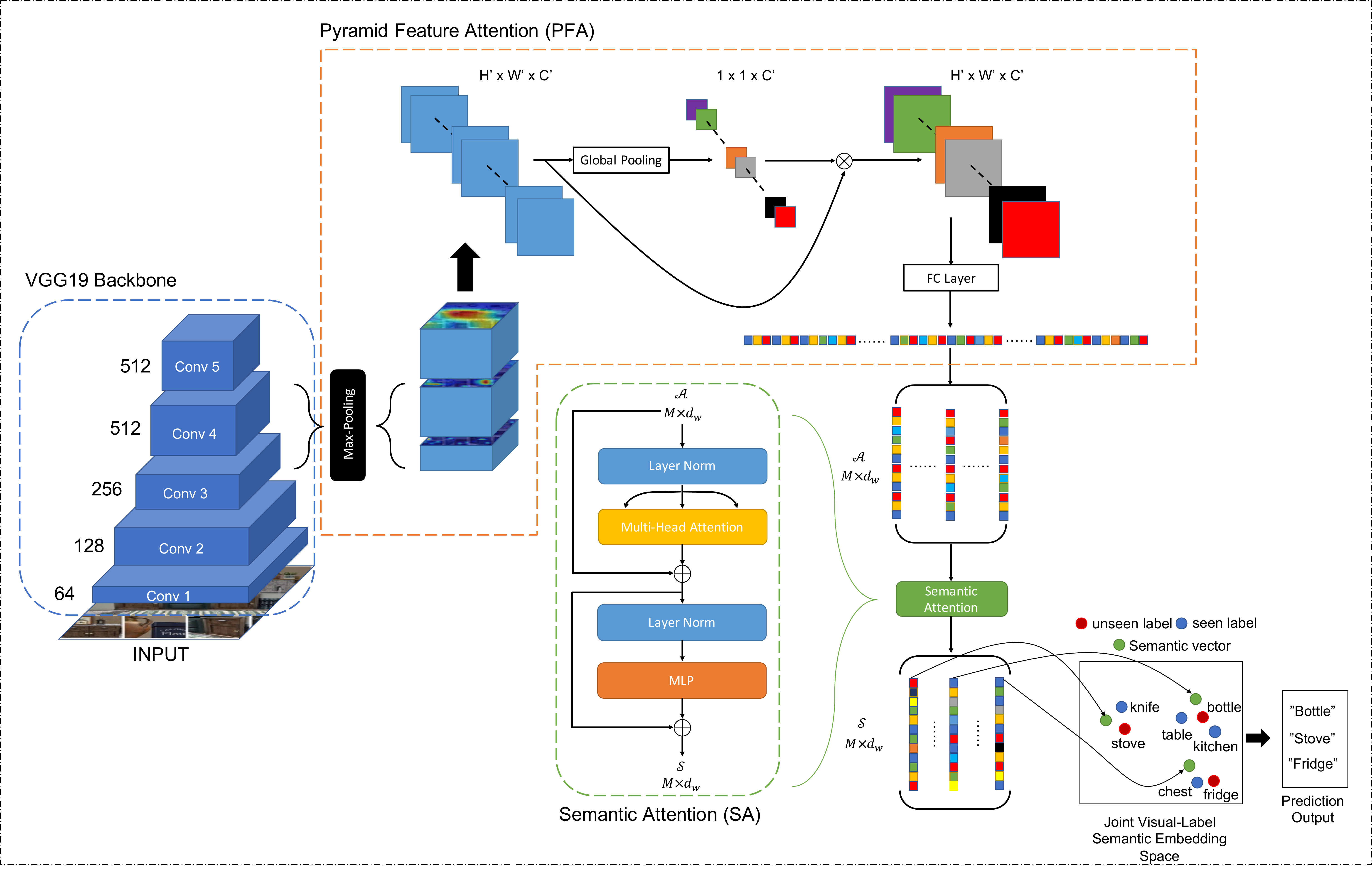}
    \caption{The overview of our proposed multi-label zero-shot learning framework. An input image is first extracted by a CNN backbone into several selected multi-scale feature maps, which are further concatenated and sent to the Pyramid Feature Attention and Semantic Attention, respectively. The obtain refined semantic vectors are then mapped into the semantic feature space to search the corresponding labels of both seen and unseen classes.
    }
    \label{fig:my_label}
\end{figure*}

\section{Proposed Method}
\subsection{Problem Setting}
We start by formalizing the task of multi-label zero-shot learning, and then introduce our proposed method in details. 
Let us denote the seen labels of a dataset as $\mathcal{C}_s$ and unseen labels as $\mathcal{C}_u$, where the seen labels are the label information contained in the image we know, that can be used in the model training process. During testing, the trained model is expected to (jointly) predict unseen labels without observing them before. Hence, the entire label information during testing is defined as $\mathcal{C} = \mathcal{C}_s\cup\mathcal{C}_u$. 
Given $n$ input samples $\left \{ \left ( I_1, Y_1\right ),\dots, \left ( I_i, Y_i\right ), \dots, \left ( I_n, Y_n\right )\right \}$ in the training process, where $I_i$ denotes the $i$-th training image and $Y_n$ denotes the corresponding set of seen labels. 
Based on the recognition setting, we construct a multi-label zero-shot learning problem that predicts the unseen labels $y_u \subset \mathcal{C}_u$ by given an image $I_u$, and a generalized multi-label zero-shot learning problem that can predict both seen and unseen labels $y_u \subset \mathcal{C}$ by given an image $I_u$, respectively. Moreover, we also follow the widely adopted assumption that each label can be represented by a semantic word vector $\left \{v^c\right \}_{c\in\mathcal{C}}$.


\subsection{Pyramid Feature Generation}

The framework of our proposed ML-ZSL method is demonstrated in Fig. 2. 
Since our method does not have any restriction on the backbone networks, so we simply use \textit{VGG19}~\cite{simonyan2014very} as an example. In view of the problem that the previous method cannot balance the major and minor classes, we believe that the reason for this imbalance is that the current multi-label zero-shot learning methods only use single high-level features. The small objects have less pixel information and are easy to be ignored during the process of sampling, so it needs to be balanced with large-scale feature maps. To solve this problem, Lin~\textit{et al.}~\cite{lin2017feature} proposed Feature Pyramid Network (FPN), which can handle the problem of multi-label object scale changes in object detection with a very small amount of calculation. On the basis of \textit{FPN}, Li~\textit{et al.}~\cite{li2017fssd} transforms the multi-layer feature maps into the same size as the bottom feature and concatenates them, then sends the concatenate feature map to the detection module, which greatly improves the detection accuracy of small objects. 

Inspired by the above two works while different from \cite{li2017fssd}, we scale the bottom feature to the same scale as the top feature and then concatenate the multi-scale feature map. The purpose of this approach is to use the sensitivity of large-scale feature maps to small objects to localize minor classes. At the same time, the small-scale feature map will make it easier to locate the major classes, and then scale the feature maps of different levels to the same size for connection, and the obtained new combined features will be able to greatly guarantee the balance between major and minor classes. As can be seen from Fig.~\ref{fig1}, there are objects of different sizes in the image, such as the smaller 'cup', the 'photo' and the larger 'stove'. These labels will all be associated with the label 'kitchen'. Therefore, we can easily find that if only the highest-level feature is used, the model cannot obtain the exact position of the 'cup', and thus cannot accurately learn the feature. Therefore, the feature maps of lower layers can play the role of localization. In addition, the cascade of the forward pass can reduce the size of the generated feature map compared to \cite{li2017fssd}, reducing the amount of computation required by the model itself. At the same time, smaller-scale feature maps are also beneficial to the extraction of semantic information. This forward feature fusion method can be expressed by the following function:
\begin{equation}
    \mathbf{F}_{fixed} = \mathcal{M}axPooling\left(\mathbf{F}_i\right),
\end{equation}
where $\mathbf{F}_i \triangleq \left[f_{i}^3, f_{i}^4\right]$ represents the feature layer $f_{i}^v$ that needs to be extracted, and $v$ represents the set of all feature layers to be extracted, usually the third and fourth different scales and their feature map sizes are $28 \times 28$ and $14 \times 14$, respectively. $i$ means the input image. $\mathcal{M}axpooling$ means the scale matching operation, we choose the max-pooling operation to highlight the local information of the bottom feature map. Then the concatenation is conducted as:
\begin{equation}
    \mathbf{C}_i = Concat\left(\mathbf{F}_{fixed}\right),
\end{equation}
where $Concat$ stands for concatenating different feature maps together to generate a new feature map $\mathbf{C}_i$.

\subsection{Pyramid Feature Attention}

After obtaining the multi-scale feature map $\mathbf{C}_i \in \mathbb{R}^{H'\times W'\times C'}$, we introduce the \textit{Pyramid Feature Attention (PFA)} to focus on the correlation among different channels. 
In this step, since the model has obtained a sufficient number of features in the previous cascade operation, the number of channels is also greatly increased. The increase of feature dimension will cause the model itself to be unable to judge the importance between different channels. Therefore, we adopt the feature channel attention mechanism to strengthen the channels with more important information and weaken the channels that are not important, thereby enhancing the learning ability of the model for important information.
We hope that the model can automatically learn the importance of different channel features, which is conducive to the subsequent generation of semantic vectors. The structure of this attention helps to enhance the network's attention to minor classes by extracting features of different scales. The \textit{PFA} is expressed by the following formula as:
\begin{equation}
    \mathbf{C}' = Map\left(\mathbf{C}_i\right),
\end{equation}
where for the input feature map $\mathbf{C}_i$, we get the new feature map $\mathbf{C}'$ after the mapping operation $Map$, and then perform global average pooling $GAPooling$ to generate channel-wise information as:
\begin{equation}
    \mathbf{z} = GAPooling\left(\mathbf{C}'\right)=\frac{1}{H'\times W'}\sum_{i=1}^{H'}\sum_{j=1}^{W'}c'(i,j).
\end{equation}

After that we can get $\mathbf{z}$, the multi-scale channel attention, next, we reduce the amount of computation by reducing the number of feature layers of the network by using the bottleneck structure. In addition, the bottleneck structure can learn nonlinear interactions between different channels, which helps to improve the correlations among channels, i.e.:
\begin{equation}
    \mathbf{z}' = \sigma[\mathbf{W}_1^{fc}\left(\mathbf{z}\right)],
\end{equation}
Where $\mathbf{W}_1^{fc}\in \mathbb{R}^{C'\times C^{ex}}$ refers to the fully convolutional layer used when $\mathbf{z}$ is channel compressed, and $\mathbf{z}'$ is the compressed attention value. The $\sigma$ represents the ReLU layer, the purpose is to increase the nonlinear information. Then restore the number of channels to $C'$:
\begin{equation}
    \boldsymbol{\alpha} = \sigma[\mathbf{W}_2^{fc}\left(\mathbf{z}'\right)],
\end{equation}
where $\mathbf{W}_2^{fc}\in \mathbb{R}^{C'\times C^{ex}}$ refers to the fully convolutional layer used to change the number of channels back to $C'$. Then perform channel-wise multiplication with $\mathbf{C}'$ to generate the PFA feature, and then perform global average pooling to get the output $\mathcal{F}$ as:
\begin{equation}
    \mathcal{F} = \boldsymbol{\alpha} \otimes \mathbf{C}' = \left[\alpha_1\left({\mathbf{c}_1}'\right)\dots\alpha_{C'}\left({\mathbf{c}_{C'}}'\right)\right].
\end{equation}

Finally, we use global average pooling to change $\mathcal{F}$ from $\mathbb{R}^{H'\times W'\times C'}$ to $\mathbb{R}^{1\times 1\times C'}$. Then we train a linear transformation to make the number of channels become $M\times d_w$, where $M$ is the number of generated semantic vectors, and $d_w$ is the dimension of the semantic vectors, usually the same length as $v^c$ (which is 300): 
\begin{equation}
    \mathcal{A} = \delta\left[GAPooling\left(\mathcal{F}\right)\right],
\end{equation}
where $\delta$ is a linear transformation. Then we get the semantic matrix $\mathcal{A} \in \mathbb{R}^{M\times d_w}$. The semantic matrix contains the semantic information of all seen labels.

\subsection{Semantic Attention}
For the generated semantic matrix $\mathcal{A} \in \mathbb{R}^{M\times d_w}$, its actual meaning is not obvious because it is only a simple segmentation of the output channels. At the same time, this segmentation operation severely destroys the global information covered by each semantic vector, thus causing troubles for our prediction. Therefore, in this section, we introduce how to add global information to the obtained semantic vectors by using our proposed \textit{Semantic Attention (SA)}.

First, we normalize the generated $\mathcal{A}$ by LayerNorm and convert it into embedding vectors $\mathbf{Q}_a$, $\mathbf{K}_a$, $\mathbf{V}_a$. 
\begin{equation}
    \mathbf{Q}_a=\mathcal{A}\mathbf{W}_a^{Q},
\end{equation}
\begin{equation}
    \mathbf{K}_a=\mathcal{A}\mathbf{W}_a^{K},
\end{equation}
\begin{equation}
    \mathbf{V}_a=\mathcal{A}\mathbf{W}_a^{V},
\end{equation}
Where $\mathbf{W}_a^{Q}$, $\mathbf{W}_a^{K}$, $\mathbf{W}_a^{V}$ are weights of which the number of channels are $d_w$. 
%
Since the generated $M$ semantic vectors are almost uncorrelated with each other at the element-wise level, we further resort to building the correlations among them by integrating a self-attention module to strengthen their coordination. Our motivation is that self-attention can reduce the dependence on external information and better capture internal correlations among features. 
To do it, we first calculate the similarity or correlation between the two based on Query $\mathbf{Q}_a$ and Key $\mathbf{K}_a$ as:
\begin{equation}
    \mathbf{r}_a = softmax(\frac{\mathbf{Q}_a\mathbf{K}_a^\top}{\sqrt{d_w}}).
\end{equation}
The second stage normalizes the above obtained scores as:
\begin{equation}
    \boldsymbol{\theta} = \mathbf{r}_a\mathbf{V}_a.
\end{equation}
After that, We add the generated $\boldsymbol{\theta}$ and $\mathcal{A}$ to a multi-layer perceptron (MLP), The purpose of MLP is to increase the non-linearity of the model while being able to perceive the local features of different vectors. Finally we get $\mathcal{S}$.
Semantic Attention is different from the traditional vision transformer~\cite{dosovitskiy2020image}. We are not dealing with different image patches, so positional encoding is not needed, and we also abandon the class/category token. We build the Multi-Head Attention operation to enrich the global information of each semantic vector. The generated new semantic matrix $\mathcal{S} \in \mathbb{R}^{M\times d_w}$ has the same dimension as $\mathcal{A}$.

\subsection{Loss Function}
During the training phase, for each input image, we output a semantic matrix $\mathcal{S} \in \mathbb{R}^{M\times d_w}$, which is used to determine the unseen class contained in the image. Therefore, inspired by~\cite{zhang2016fast}, when calculating class similarity, we need to provide a higher ranking for the positive class, and a lower-ranking for the negative class, as the basis for the design of our loss function: 
\begin{equation}
    \mu_{jk}=\max\left(\mathcal{S}n_j\right) - \max\left(\mathcal{S}p_k\right),
\end{equation}
where $n_j$ is the word vector of the negative class, and $p_k$ is the word vector of the positive class. We choose the $\max$ result for calculation, the purpose is to make all classes output high scores while contributing to semantic diversity. Similar ideas are also used in multi-label support vector machines (SVM)~\cite{crammer2001algorithmic}. 
Next, inspired by~\cite{zhang2016fast}, we design the following modified ranking loss as:
\begin{equation}
    \mathcal{L}_{rank} = \beta\sum_j\sum_k\log \left(1+e^{\mu_{jk}}\right),
\end{equation}
where $\beta=\left(\left|Y\right|\left|\bar{Y}\right|\right)^{-1}$. $\left|Y\right|$ and $\left|\bar{Y}\right|$ denote the number of seen and unseen classes, respectively. The hyper-parameter $\beta$ is used to normalize the ranking loss.

In addition, in order to reduce the difficulty of classifying images with high label diversity, 
we propose to add a new weight to the ranking loss as:
\begin{equation}
    \omega = 1 + \sum_i var(Y_i). 
\end{equation}
Moreover, in order to balance the semantic diversity between different vectors and encourage the network to learn relevant information between different labels, we further construct a regularization loss as:
\begin{equation}
    \mathcal{L}_{reg} = \left \| \sum_n var(\mathcal{S}_n) \right \|_1.
\end{equation}

In summary, our final loss function for each image is defined as:
\begin{equation}
    \mathcal{L}_{final} = \frac{1}{N}\sum_{i=1}^N\left(w \cdot (1-\lambda)\mathcal{L}_{rank}(\mathcal{S}_i, Y_i) + \lambda \mathcal{L}_{reg}(\mathcal{S}_i)\right),
\end{equation}
where $N$ is the value of batch size, and $\lambda$ is a hyper-parameter that regularize the weights.
 
\section{Experiments}
\subsection{Experimental Setup}

\subsubsection{Datasets}
First, we evaluate our experiments on the \textit{NUS-WIDE}~\cite{chua2009nus}.
The \textit{NUS-WIDE} dataset contains 270k images which are annotated by human into 81 classes. These 81 classes are also called 'ground-truth' labels. Besides, each image also has 925 labels which are extracted from Flickr user tags. Similar to the previous works~\cite{huynh2020shared, zhang2016fast}, we choose the 925 labels as the seen labels, and the other 81 human-annotated labels as unseen. 
Next, We also test the ZSL and GZSL tasks separately on the larger multi-label dataset \textit{Open-Images}-V4~\cite{kuznetsova2020open}, which contains nearly 9 million training images and 125,436 test images, as well as 41,620 images as 
the validation set. The training set contains 7186 visible labels, and each label contains at least 100 samples for training. In the test set, we select the top 400 most frequent classes not present in the training data, each unseen label has at least 75 samples for detection. Due to the large number of classes, each image has at least one unannotated label.

\subsubsection{Evaluation Metrics}
In order to measure the performance of our proposed model under the multi-label zero-shot classification task, we use the mean Average Precision (mAP) and F1 score at \textit{top-K} predictions to evaluate. The \textit{top-K} F1 score is used to measure the accuracy of the model for predicting the labeling of the image, while the mAP measurement is based on the accuracy of label retrieval, showing the model's ranking accuracy for each label of the image.

\subsubsection{Implementation Details}
Unlike \cite{huynh2020shared} and \cite{zhang2016fast}, this is an end-to-end training model. We use the backbone network \textit{VGG19} pre-trained on the \textit{ImageNet} dataset \cite{deng2009imagenet}. During the cascade operation of multi-scale feature maps, we extract feature maps with sizes of $28\times28$, $14\times14$ and $7\times7$ respectively, and use max-pooling to directly sample the bottom feature maps to $7\times7$. Compared with \textit{FSSD}~\cite{li2017fssd}, this operation greatly reduces the amount of calculation. The feature map size for pyramid feature attention is $2048\times7\times7$. 
We choose the Adam optimizer~\cite{kingma2014adam} for the model training, and the maximum learning rate is set as $1e^{-5}$, which is reduced to $\frac{1}{10}$ of the previous after 5 epochs. The weight decay is $4e^{-3}$. In the loss function, the value of the regularization term $\lambda$ is 0.4, and the number $M$ of semantic vectors generated by the model is 8. The number of epochs for experiments in the NUS-WIDE dataset is 10, and batch\_size is set to 64. 

\subsubsection{Baselines}
We compare our method with several representative ML-ZSL methods including 
\textit{CONSE}~\cite{norouzi2014zero}, 
\textit{LabelEM}~\cite{akata2015label}, 
\textit{Fast0Tag}~\cite{zhang2016fast}, 
Kim~\textit{et al.}~\cite{kim2018bilinear}, 
\textit{LESA}~\cite{huynh2020shared}, 
and Gupta~\textit{et al.}~\cite{gupta2021generative}. 
The selection of our competitors are based on the following criteria: 1) All competitors are published or presented in the most recent years; 2) A wide range of techniques are covered; 3) They have stated or demonstrated the state-of-the-art in ML-ZSL; and 4) All competitors are evaluated fairly under the same criteria with their best (reported) results. 
Moreover, we also measure and analyze our method. Specifically, we not only adjust the number of $M$ but also choose the best regularization term and study the influence of each attention module by using the model's experimental results.

\begin{table*}[htbp]
  \centering
  \caption{State-of-the-art comparison for multi-label ZSL and GZSL tasks on the \textbf{NUS-WIDE} dataset in detail. We show the indicators of F1-Score in the case of $K\in{3, 5}$ and mAP. We can clearly see that our method has surpassed all state-of-the-art methods, whether it is under ZSL or GZSL conditions. Best results are shown in bold.}
    \scalebox{0.95}{
    \begin{tabular}{lc|cccccc|c}
    \toprule
    \textbf{Method} & \textbf{Task} & \textbf{P (K = 3)} & \textbf{R (K = 3)}  & \textbf{F1 (K = 3)} & \textbf{P (K = 5)} & \textbf{R (K = 5)} & \textbf{F1 (K = 5)} & \textbf{mAP} \\
    \midrule
    \multirow{2}[2]{*}{\textit{CONSE}~\cite{norouzi2014zero}}  & ZSL   & 17.5 & 28.0   & 21.6  & 13.9 & 37.0 & 20.2 & 9.4\\
                               & GZSL  & 11.5 & 5.1     & 7.0   & 9.6 & 7.1  & 8.1  & 2.1\\
    \cmidrule{2-9}  
    \multirow{2}[2]{*}{\textit{LabelEM}~\cite{akata2015label}}& ZSL   & 15.6  & 25.0  & 19.2 & 13.4  & 35.7  & 19.5 & 7.1\\
                               & GZSL  & 15.5  & 6.8  & 9.5 & 13.4  & 9.8  & 11.3 & 2.2\\
    \cmidrule{2-9}  
    \multirow{2}[2]{*}{\textit{Fast0Tag}~\cite{zhang2016fast}}& ZSL  & 22.6  & 36.2  & 27.8 & 18.2  & 48.4  & 26.4 & 15.1 \\
                                & GZSL & 18.8  & 8.3  & 11.5 & 15.9  & 11.7  & 13.5 & 3.7 \\
    \cmidrule{2-9}  
    \multirow{2}[2]{*}{Attention per Label~\cite{kim2018bilinear}}& ZSL   & 20.9  & 33.5  & 25.8 & 16.2  & 43.2  & 23.6 & 10.4 \\
                                           & GZSL  & 17.9  & 7.9   & 10.9 & 15.6  & 11.5  & 13.2 & 3.7 \\
    \cmidrule{2-9}  
    \multirow{2}[2]{*}{Attention per Cluster~\cite{huynh2020shared}} & ZSL   & 20.0  & 31.9  & 24.6 & 15.7  & 41.9  & 22.9 & 12.9 \\
                                              & GZSL  & 10.4  & 4.6  & 6.4 & 9.1  & 6.7  & 7.7 & 2.6 \\
    \cmidrule{2-9}  
    \multirow{2}[2]{*}{\textit{LESA} (M = 10)~\cite{huynh2020shared}} & ZSL   & 25.7  & 41.1  & 31.6 & 19.7  & 52.5  & 28.7 & 19.4 \\
                                      & GZSL  & 23.6  & 10.4  & 14.4 & 19.8  & 14.6  & 16.8 & 5.6 \\
    \cmidrule{2-9}
    \multirow{2}[2]{*}{Gupta~\textit{et al.}~\cite{gupta2021generative}} & ZSL   & 26.6  & \textbf{42.8}  & 32.8 & 20.1  & 53.6  & 29.3 & 25.7 \\
                                      & GZSL  & 30.9  & 13.6  & 18.9 & 26.0  & 19.1  & 22.0 & 8.9 \\
    \cmidrule{2-9}    
    \multirow{2}[2]{*}{\bf{Our Approach}}& ZSL   & \bf{34.0}     & 42.3     & \bf{37.7} & \bf{26.7}     & \bf{55.3}     & \bf{36.0} & \bf{28.0}\\
                                    & GZSL  & \bf{31.2}     & \bf{13.9}     & \bf{19.2} & \bf{26.4}     & \bf{19.6}     & \bf{22.5} & \bf{9.3} \\
    \bottomrule
    \end{tabular}}%
  \label{tab1}%
\end{table*}

\begin{table*}[htbp]
  \centering
  \caption{State-of-the-art comparison for multi-label ZSL and GZSL tasks on the \textbf{Open-Images} dataset in detail. We show the indicators of F1-Score in the case of $K\in{10, 20}$ and mAP. We can clearly see that our method has surpassed all state-of-the-art methods, whether it is under ZSL or GZSL conditions. Best results are shown in bold.}
    \scalebox{0.95}{
    \begin{tabular}{lc|cccccc|c}
    \toprule
    \textbf{Method} & \textbf{Task} & \textbf{P (K = 10)} & \textbf{R (K = 10)}  & \textbf{F1 (K = 10)} & \textbf{P (K = 20)} & \textbf{R (K = 20)} & \textbf{F1 (K = 20)} & \textbf{mAP} \\
    \midrule
    \multirow{2}[2]{*}{\textit{CONSE}~\cite{norouzi2014zero}}  & ZSL   & 0.2 & 7.3   & 0.4  & 0.2 & 11.3 & 0.3 & 40.4\\
                               & GZSL  & 2.4 & 2.8     & 2.6   & 1.7 & 3.9  & 2.4  & 43.5\\
    \cmidrule{2-9}  
    \multirow{2}[2]{*}{\textit{LabelEM}~\cite{akata2015label}}& ZSL   & 0.2  & 8.7  & 0.5 & 0.2  & 15.8  & 0.4 & 40.5\\
                               & GZSL  & 4.8  & 5.6  & 5.2 & 3.7  & 8.5  & 5.1 & 45.2\\
    \cmidrule{2-9}  
    \multirow{2}[2]{*}{\textit{Fast0Tag}~\cite{zhang2016fast}}& ZSL  & 0.3  & 12.6  & 0.7 & 0.3  & 21.3  & 0.6 & 41.2 \\
                                & GZSL & 14.8  & 17.3  & 16.0 & 9.3  & 21.5  & 12.9 & 45.2 \\
    \cmidrule{2-9}  
    \multirow{2}[2]{*}{Attention per Cluster~\cite{huynh2020shared}} & ZSL   & 0.6  & 22.9  & 1.2 & 0.4  & 32.4  & 0.9 & 40.7 \\
                                              & GZSL  & 15.7  & 18.3  & 16.9  & 9.6  & 22.4  & 13.5 & 44.9 \\
    \cmidrule{2-9}  
    \multirow{2}[2]{*}{\textit{LESA} (M = 10)~\cite{huynh2020shared}} & ZSL   & 0.7  & 25.6  & 1.4  & 0.5  & 37.4  & 1.0 & 41.7 \\
                                      & GZSL  & 16.2  & 18.9  & 17.4 & 10.2  & 23.9  & 14.3 & 45.4 \\
    \cmidrule{2-9}  
    \multirow{2}[2]{*}{Gupta~\textit{et al.}~\cite{gupta2021generative}} & ZSL   & 1.3  & 42.4  & 2.5 & \textbf{1.1}  & 52.1  & \textbf{2.2} & 43.0 \\
                                      & GZSL  & 33.6  & 38.9  & 36.1 & 22.8  & 52.8  & 31.9 & 49.7 \\
    \cmidrule{2-9} 
    \multirow{2}[2]{*}{\bf{Our Approach}}& ZSL   & \bf{1.5}     & \bf{49.9}     & \bf{2.9} & \bf{1.1}     & \bf{64.0}     & \bf{2.2} & \bf{45.4}\\
                                    & GZSL  & \bf{33.9}     & \bf{40.1}     & \bf{36.7} & \bf{23.1}     & \bf{53.1}     & \bf{32.2} & \bf{51.5} \\
    \bottomrule
    \end{tabular}}%
  \label{tab2}%
\end{table*}

\subsection{State-of-the-art Comparison}

\subsubsection{Result on NUS-WIDE Dataset}
Table~\ref{tab1} shows the comparison between our model and other current state-of-the-art methods on the \textit{NUS-WIDE} dataset, Not only the traditional zero-shot learning (ZSL), but also the results of generalized zero-shot learning (GZSL) are also reported. Due to the limitation of space, we only show the most important \textit{Top-K} $(K=3, 5)$ F1 Score and mAP results. 

For the ZSL task, the performance of \textit{CONSE}~\cite{norouzi2014zero} and \textit{LabelEM}~\cite{akata2015label} on the \textit{NUS-WIDE} dataset, whether it is mAP or F1-Score, cannot be compared with the current methods. Because \textit{Fast0Tag}~\cite{zhang2016fast} learns to choose the main direction of the image in the word vector space, the mAP is improved by 5.7\% compared to CONSE. For the newly proposed method \textit{LESA}~\cite{huynh2020shared}, which introduces a spatial attention mechanism, predicts unseen labels by splitting the image into patches and sharing the attention between different them. It also achieved the best results before. But our approach is now the best, leading \textit{LESA} by 8.6\% in mAP and at least 6.1\% improvement in F1-Score. The overall effect is greatly improved. In the comparison with Gupta~\textit{et al.}~\cite{gupta2021generative}, it is difficult to judge the performance from the methodology due to the differences between us. However, from the experimental performance on the \textit{NUS-WIDE} test set, our method outperforms the Gupta~\textit{et al.}~\cite{gupta2021generative} in 6 out of 7 metrics. Especially the performance of the most important F1-Score and mAP. On the ZSL task, our method is generally higher than 5\% in the F1-Score, while the mAP result is also 2.3\% higher.

Under the conditions of GZSL, due to the sharp increase in the number of labels that need to be predicted, the mAP of all methods has dropped drastically. But \textit{LESA} is still the best performer among other attention-based comparison methods, reaching 5.6\%. However, our approach reached 9.3\%, which exceeded 3.7\% to \textit{LESA}. This benefits from the more global information contained in semantic vectors, which makes our prediction results closer to the image feature. In subsequent comparisons with Generative ZSL, i.e., Gupta~\textit{et al.}~\cite{gupta2021generative}, our method achieves the best results in all evaluation metrics. In GZSL, as the task becomes more difficult, the gap between the various methods is not particularly obvious. In the mAP evaluation metric, our method also achieved the best results, which was 0.4\% higher than that of Gupta~\textit{et al.} \cite{gupta2021generative}.

\subsubsection{Results on Open-Images Dataset}
Table~\ref{tab2} shows the performance of our method and all current state-of-the-art comparison methods on the \textit{Open-Images} dataset. Like \textit{NUS-WIDE}, we compare on two tasks, ZSL and GZSL. We show the most important \textit{Top-K} $(K=10, 20)$ F1 Score and mAP results. Each image in \textit{Open-Images} has at least one class, and because the total number of classes is too large, the semantic space will also become more complex, making model prediction more difficult.

On the ZSL task, all methods perform poorly in precision. The mAP results of \textit{CONSE}~\cite{norouzi2014zero} and \textit{LabelEM}~\cite{akata2015label} are only 0.2\%, while other attention-based comparison models, such as \textit{Fast0Tag}~\cite{zhang2016fast} and  \textit{LESA}~\cite{huynh2020shared} are not more than 1\%. And Gupta~\textit{et al.}~\cite{gupta2021generative} based on generative models is the only one that exceeds 1\%. However, our method achieves the best results at the precision of $K=10 and 20$. In addition, the recall of our model has also been greatly improved compared to Gupta~\textit{et al.}~\cite{gupta2021generative}. The final mAP also proves the effectiveness of our method on the largest dataset from another dimension.

In the GZSL task, due to the huge increase in the number of classes in the test session, the number of predictable labels in each image is increased by at least 40 times. Therefore, the numerical results of all methods are greatly improved in the case of $K=10, 20$. Compared with the current state-of-the-art attention-based model \textit{LESA}~\cite{huynh2020shared}, our method is far ahead in terms of results. The lead on the F1-Score metric is close to 20\%, while it also has a 6.6\% lead on mAP. At the same time, our model also leads by a certain degree compared to Gupta~\textit{et al.}~\cite{gupta2021generative}.

\subsection{Effect of Hyper-parameters}
\begin{figure*}
    \centering
    \subfigure[The value of F1-Score (K=3) when the regularization weight $(\lambda)$ changes from 0 to 1. ]{
    \includegraphics[width=0.31\textwidth]{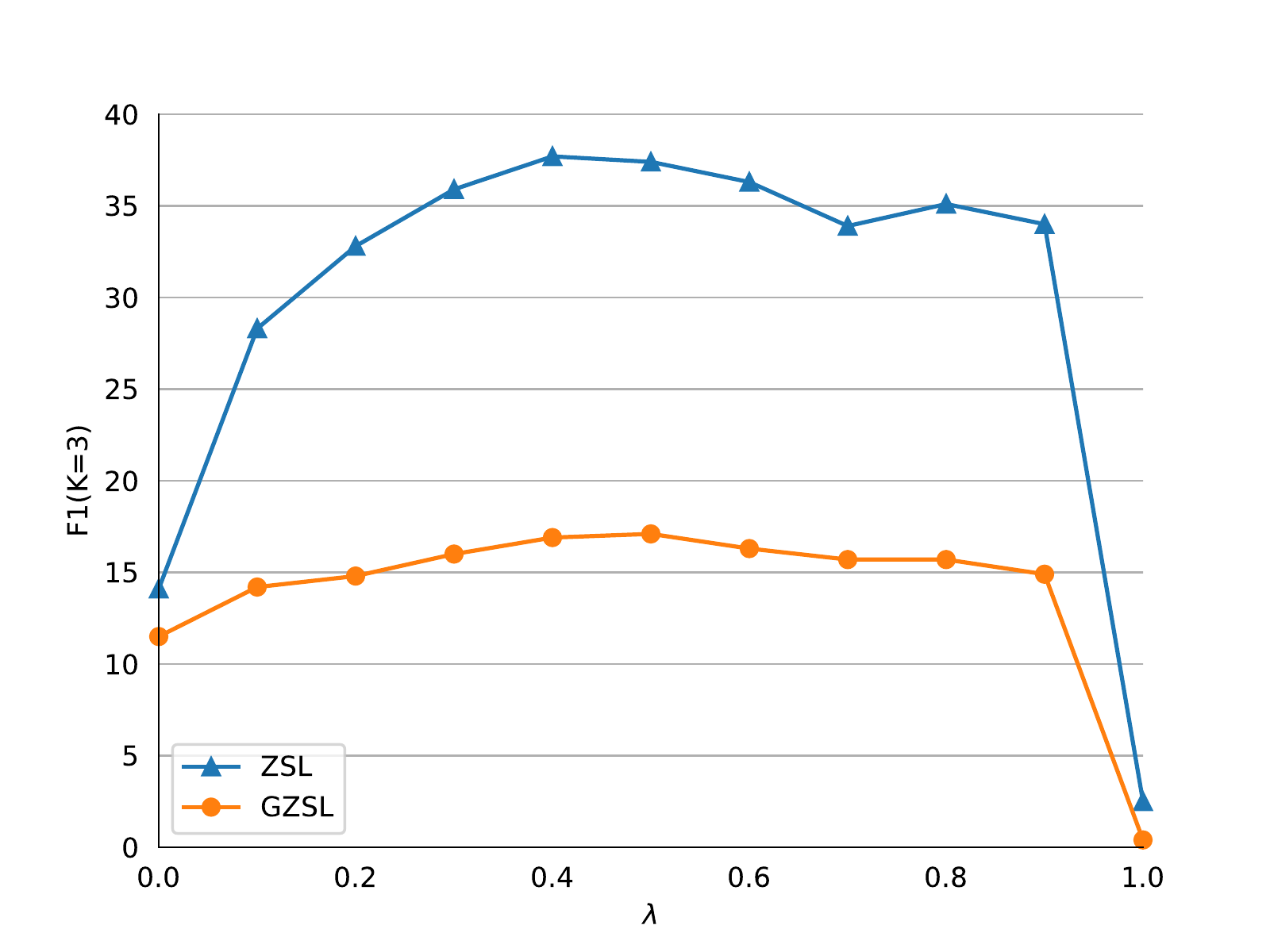}}
    \hspace{0.2cm}
    \subfigure[The value of mAP when the regularization weight $(\lambda)$ changes from 0 to 1.]{
    \includegraphics[width=0.31\textwidth]{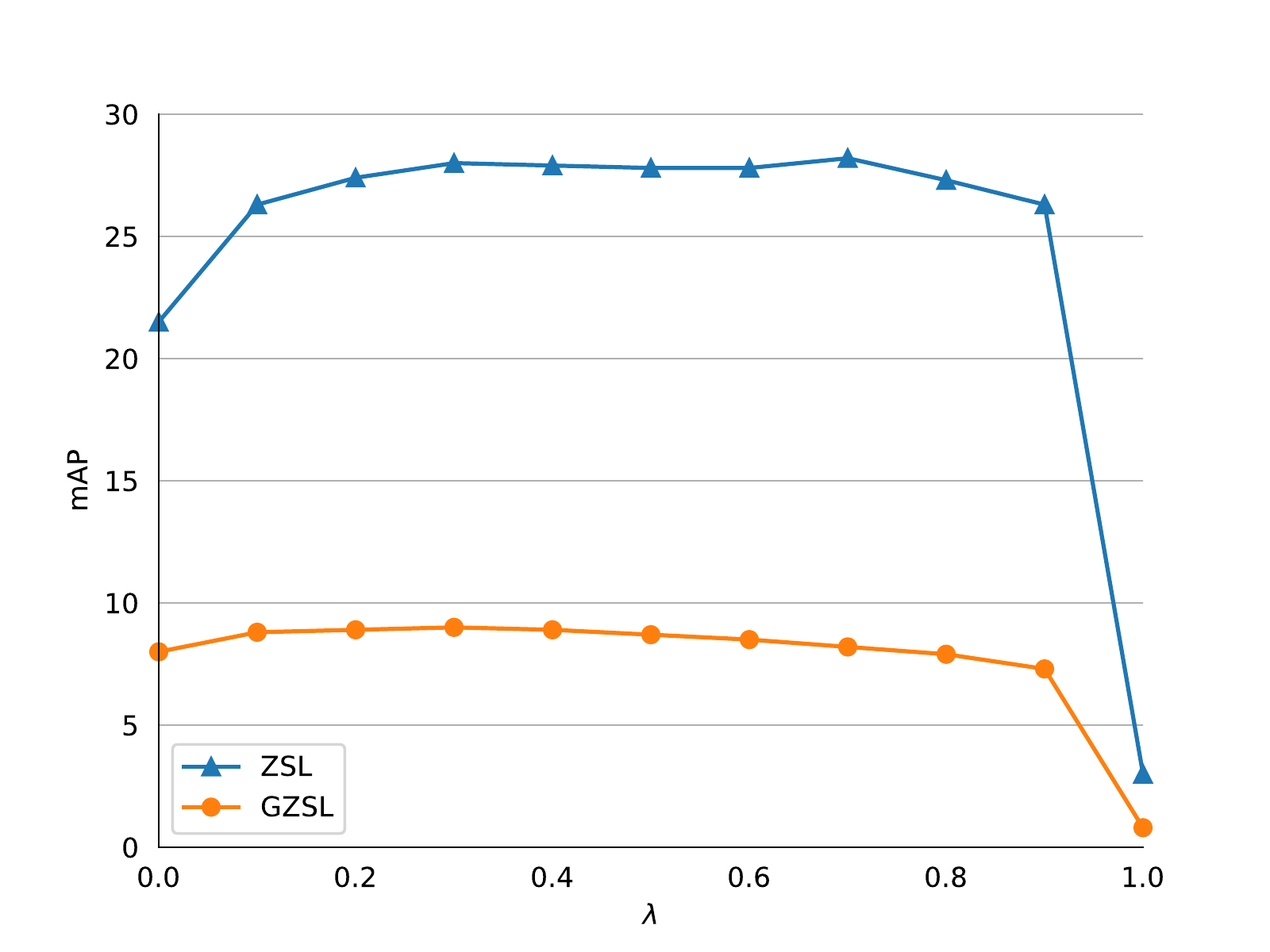}}
    \hspace{0.2cm}
    \subfigure[The value of mAP when the number of generated semantic vectors $(M)$ changes from 1 to 10.]{
    \includegraphics[width=0.31\textwidth]{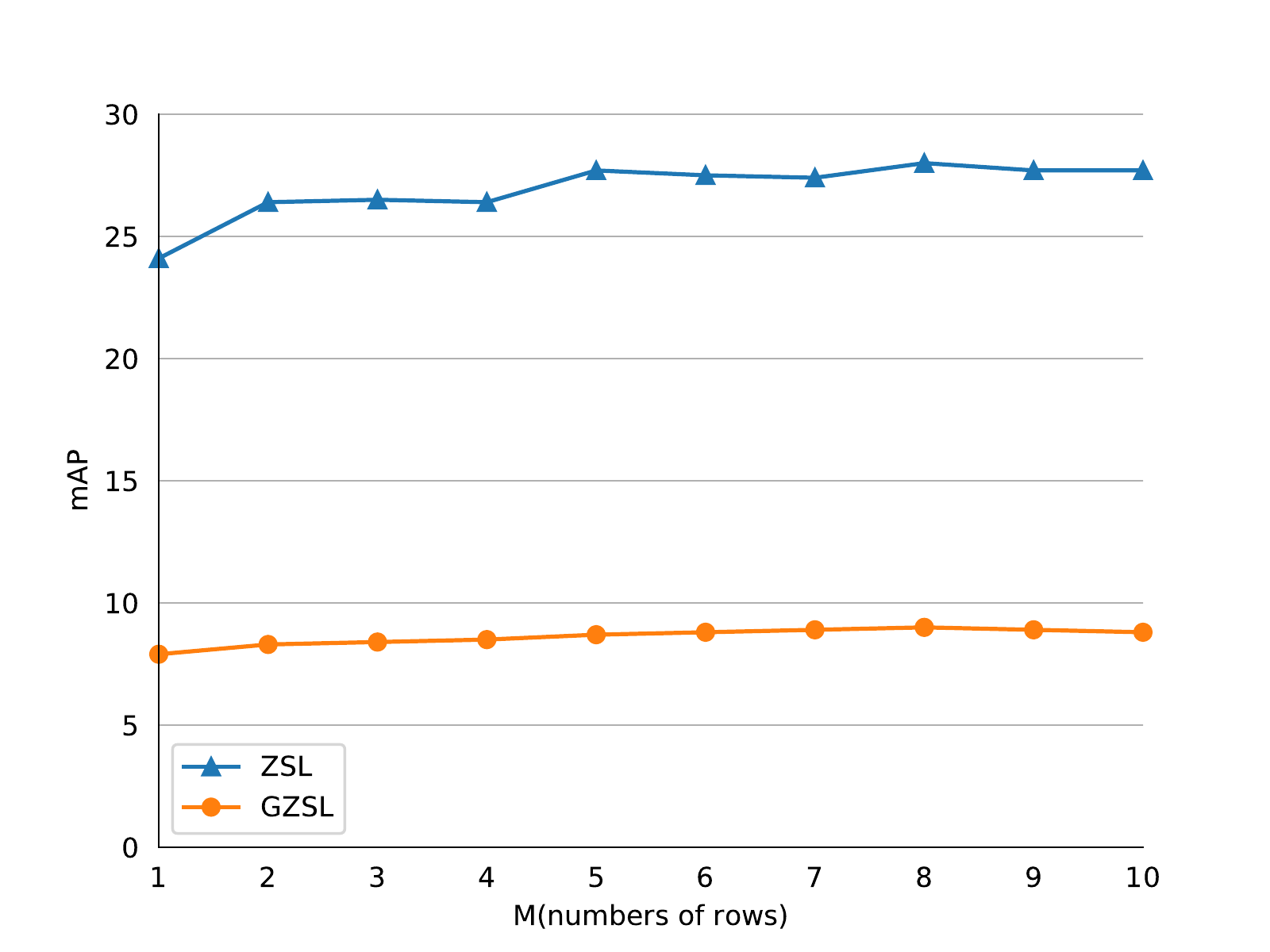}}
\caption{An illustration of the impact of choosing different hyper-parameters for ZSL and GZSL tasks. All ZSL and GZSL tasks are performed on the \textit{NUS-WIDE} test set.}
\label{fig:hyperparameter}
\end{figure*}

Fig.~\ref{fig:hyperparameter}(a), Fig.~\ref{fig:hyperparameter}(b), and Fig.~\ref{fig:hyperparameter}(c) demonstrate the effect of hyper-parameters in our method. 
Specifically, there are two hyper-parameters in our method, namely, $M$ and $\lambda$, where $M$ is the number of generated semantic vectors and its number determines the breadth of the model's prediction. Theoretically, the larger $M$ is, the more abundant types of predicted labels will be generated. However, since the amount of information contained in a single image is limited, an increase in $M$ will obscure the information contained in each semantic vector. 


From Fig.~\ref{fig:hyperparameter}(c), we can conclude that in the ZSL curve, when the value of $M$ increases from 1 to 10, it will first increase the value of mAP, and reach the highest value when $M=8$. Then as the value of $M$ increases, the mAP begins to fall gently. The GZSL curve is relatively stable than the ZSL curve. It also reaches its peak at $M=8$, and then drops slightly, but the overall curve is very stable. This phenomenon also better confirms our previous assumptions about the hyper-parameter $M$. Therefore, our model outputs the best result when $M=8$. At the same time, the change of $M$ will not greatly affect the accuracy of the model itself. Compared with \textit{LESA}, when $M=1$ and $M=10$, the changes brought by mAP are significant.

Next is the hyper-parameter $\lambda$. As a regularization weight, $\lambda$ is very important for the loss function. A good and stable loss function is the 'teacher' who guides the model to perform well during training process. 
First we can see from Fig.~\ref{fig:hyperparameter}(a) that the effect of the change of $\lambda$ on F1-Score is very obvious. Especially when $\lambda=0$ and $\lambda=1$. When $\lambda=0$, the loss function of the model loses the regularization term, resulting in the lack of constraints of the generated semantic vectors, resulting in a decrease in accuracy. When $\lambda=1$, the loss function only contains the regularization term, which is meaningless to the model's loss, so the F1-Score approaches 0. 
Fig.~\ref{fig:hyperparameter}(b) shows the effect of changes in $\lambda$ on the results of mAP. We can clearly see through that, when $\lambda=0.4$, the model achieved the best mAP and F1-Score. Except for the two extreme conditions of $\lambda=0$ and $\lambda=1$, when $\lambda$ becomes other values, the change in mAP is not obvious. Therefore, the change of $\lambda$ will not affect the model's ability to recognize labels, but will affect the model's ability for prediction.

\subsection{Ablation Study}
To better understand the contribution of each component in our method and measure their importance, we demonstrate the ablation study of our method in TABLE~\ref{tab4}. In this results, 'None' denotes the scenario that we only use \textit{VGG19} to generate semantic vectors for unseen classes. It can be seen that after the model uses the pyramid structure, the prediction ability in the ZSL task has been improved by 1.6\%, and the GZSL has also been improved by 0.8\%, which shows that the rational use of features at different levels has greatly exceeded the efficiency. At the same time, it also proves the validity of our judgment on the existence of major and minor classes. In the use of \textit{PFA} structure and \textit{SA} structure, based on the pyramid feature, both \textit{PFA} and \textit{SA} can play a positive role in the prediction ability of the model. However, compared with \textit{SA}, \textit{PFA} is better matched with pyramid feature. When the two act on the model at the same time, the effect is improved by 1.4\% on the ZSL task and 0.7\% on the GZSL task compared with only \textit{PFA}. The change of $M$ and regularization weight in the ablation experiment is to reflect whether different hyper-parameters will have an impact on the prediction results of the model for unseen classes. The \textit{PFA} and \textit{SA} have significantly improved the capabilities of the model, and finally achieved the best results when the model reached the most ideal state.

\begin{table}[h]\normalsize
\centering
\caption{Ablation study shows the contribution of the different components in our proposed approach, and also shows the results of original implementation. The baseline methods are performed on the \textit{NUS-WIDE} test set.}
\begin{tabular}{c|cccc}
                                &                    &     & \multicolumn{2}{c}{mAP} \\ \cline{4-5} 
Module                          & M                  & Reg & ZSL        & GZSL       \\ \hline
None                            & 8                  & 0.4 & 24.3       & 7.4        \\ \hline
Pyramid                         & 8                  & 0.4 & 25.9       & 8.2        \\ \hline
Pyramid+PFA                     & 8                  & 0.4 & 26.6       & 8.6        \\ \hline
Pyramid+SA                      & 8                  & 0.4 & 26.3       & 8.4        \\ \hline
\multirow{3}{*}{Pyramid+PFA+SA} & 3                  & 0.4 & 27.1       & 8.6        \\ \cline{2-5} 
                                & \multirow{2}{*}{8} & 0.4 & \textbf{28.0}       & \textbf{9.3}        \\ \cline{3-5} 
                                &                    & 0.8 & 27.5       & 8.8       
\end{tabular}

\label{tab4}
\end{table}



\subsection{Multi-label Learning}
In this section, in order to explore the performance of this model in the traditional multi-label learning problem, we design a new experiment. Each label contains training images, but the number of samples among labels is very different. Our comparison methods include Logistic Regression \cite{tsoumakas2007multi}, \textit{WSABIE} \cite{weston2011wsabie}, \textit{WARP} \cite{gong2013deep}, \textit{Fast0Tag} \cite{zhang2016fast}, \textit{CNN-RNN} \cite{wang2016cnn}, One Attention per Label using Bilinear Attention Network~\cite{kim2018bilinear}, and \textit{LESA} \cite{huynh2020shared}. 

TABLE~\ref{tab3} shows the F1-Score at $K\in\{3,5\}$ and mAP values at 81 `ground truth' labels on \textit{NUS-WIDE}. Obviously, compared with the multi-label zero-shot learning method \textit{Fast0Tag}~\cite{zhang2016fast} and \textit{LESA}~\cite{huynh2020shared}, our mAP value is 13.6\% higher than that of \textit{LESA}~\cite{huynh2020shared} on the \textit{NUS-WIDE} dataset, becoming the best performing multi-label zero-shot learning method. In addition, our mAP is also 12.5\% higher than the second place One Attention per Label~\cite{kim2018bilinear}, which is the highest mAP among all comparison methods. The success of this experiment shows that our model not only has strong predictive ability, but also has strong competitiveness in the field of classification and recognition.

\begin{table*}[htbp]
  \centering
  \caption{Performance of \textbf{Multi-label Learning} methods on \textbf{NUS-WIDE} dataset in detail. It can be seen that our method has achieved significant results. The best results are in bold. }
    \scalebox{1}{
    \begin{tabular}{lcccccc|c}
    \toprule
    \textbf{Method}     & \textbf{P(K=3)}  & \textbf{R(K=3)} & \textbf{F1(K=3)} & \textbf{P(K=5)}  & \textbf{R(K=5)} & \textbf{F1(K=5)} &\textbf{mAP} \\
    \midrule
    \textit{Logistic}~\cite{tsoumakas2007multi}                      & 46.1   & 57.3   & 51.1  & 34.2  & 70.8  & 46.1  & 21.6 \\
    \textit{WARP}~\cite{gong2013deep}                                & 49.1   & 61.0   & 54.4  & 36.6  & 75.9  & 49.4  & 3.1 \\
    \textit{WSABIE}~\cite{weston2011wsabie}                          & 48.5   & 60.4   & 53.8  & 36.5  & 75.6  & 49.2  & 3.1 \\
    \textit{Fast0Tag}~\cite{zhang2016fast}                           & 48.6   & 60.4   & 53.8  & 36.0  & 74.6  & 48.6  & 22.4 \\
    \textit{CNN-RNN}~\cite{wang2016cnn}                              & 49.9   & 61.7   & 55.2  & 37.7  & 78.1  & 50.8  & 28.3 \\
    One Attention per Label~\cite{kim2018bilinear}          & 51.3   & 63.7   & 56.8  & 38.0  & 78.8  & 51.3  & 32.6 \\
    One Attention per Cluster (M = 10)~\cite{huynh2020shared}  & 51.1   & 63.5   & 56.6   & 37.6   & 77.9   & 50.7  & 31.7 \\
    \textit{LESA} (M = 1)~\cite{huynh2020shared}                        & 51.4   & 63.9   & 57.0   & 37.9   & 78.6   & 51.2  & 29.6 \\
    \textit{LESA} (M = 10)~\cite{huynh2020shared}                       & 52.3   & 65.1   & 58.0   & 38.6   & 80.0   & 52.0  & 31.5 \\
    \bf{Our Approach}                                & \bf{52.7}& \bf{65.5}& \bf{58.4}     & \bf{39.0}& \bf{80.8}& \bf{52.6}     & \bf{45.1} \\
    \bottomrule
    \end{tabular}}%
  \label{tab3}%
\end{table*}

\begin{figure*}[htbp]
    \centering
    \includegraphics[width=0.68\textwidth]{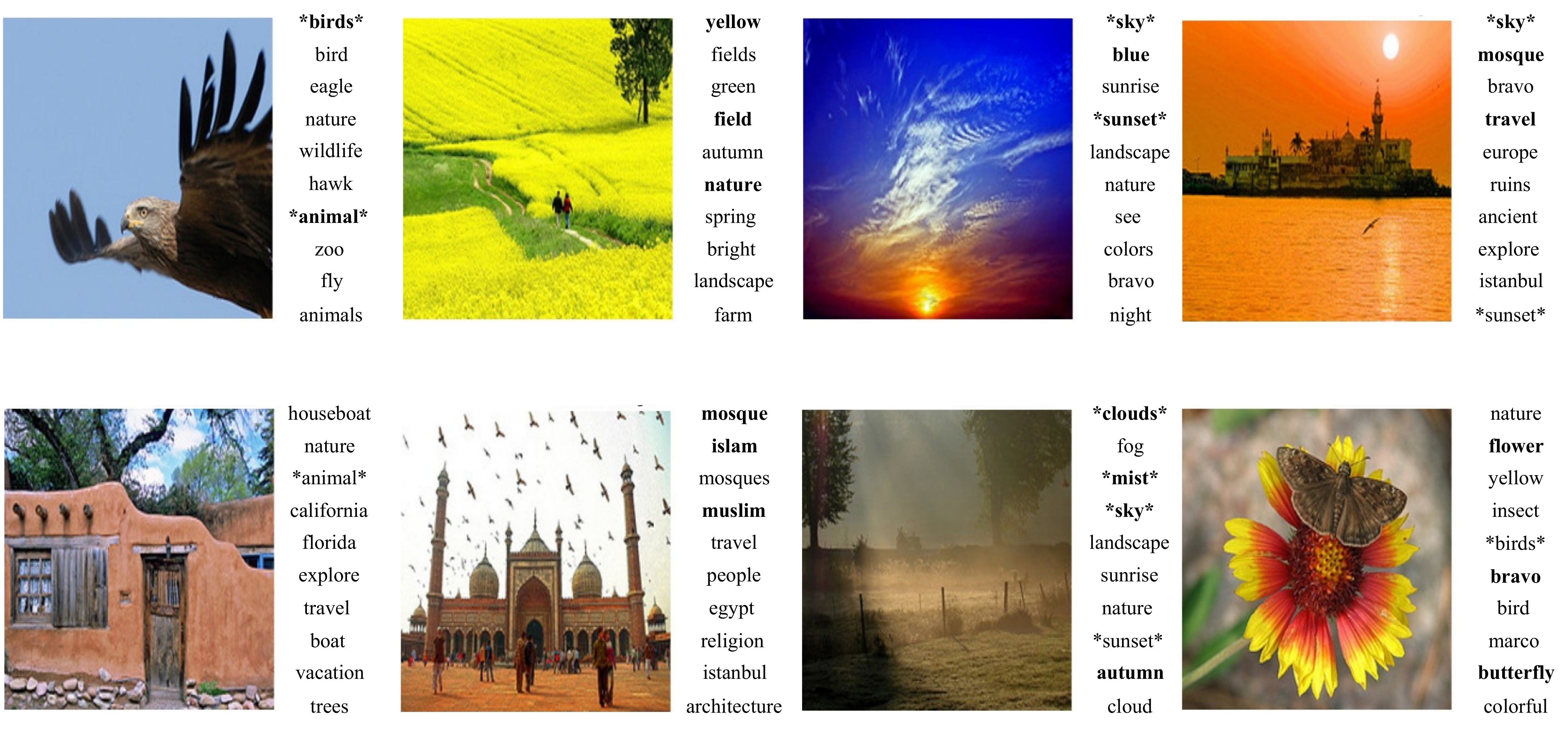}
    \caption{Qualitative results of our method, which shows the top-10 predicted labels. Asterisks mark the unseen labels, while bold text represents the correct labels according to the provided ground truth in \textit{NUS-WIDE} test set. }
    \label{fig3}
\end{figure*}

\subsection{Qualitative Results}
In this paper, our method generates a set of semantic vectors that represent many different principal orientations of test images in the visual label semantic space. These vectors are calculated with the semantics in the space, and the closest top 10 classes are obtained as prediction results, so that our model has the ability to predict seen and unseen labels. These labels include not only nouns, but also abstract labels that are difficult to classify in recognition tasks, such as adjectives and verbs. For example, in the first picture, we can see that we have successfully predicted 'bird' and 'animal'. In addition, our model has also predicted 'eagle', and combined with sky features, we can also predict verb descriptions such as 'fly'. Similarly, the third picture is also very good at identifying the sunset, indicating that the network has a strong ability to perceive the environment. The sixth picture, for the judgment of the building, is a test of the mastery of the details of the image. Our method is able to accurately identify building classes with high accuracy. In the seventh image, our model is able to make good use of the smaller feature maps to obtain blur information to recognize fog. In the last image, the model accurately predicted 'flower' and 'butterfly' while also being able to associate them with the adjective 'bravo'. It shows that our model can accurately find the best matching classes in the semantic space when generating semantic vectors. In summary, we use \textit{PFA} to balance the major and minor classes, and \textit{SA} can establish a relationship with other generated vectors while ensuring that the generated semantic vectors retain their original information. This enables each individual semantic vector with more accurate approximation on the predicted classes in the semantic feature space.


\section{Conclusion}
In this paper, we proposed a novel unbiased multi-label zero-shot learning framework involving the proposed Pyramid Feature Attention and Semantic Attention, to jointly address the problem of imbalance between major classes and minor classes by extracting the attention between multi-scale feature layers. 
Extensive experiments on the large-scale \textit{NUS-WIDE} and \textit{Open-Images} datasets show that our framework has achieved state-of-the-art results and significantly improved the classification accuracy in the multi-label environment.

\bibliographystyle{IEEEtran}
\bibliography{mlzsl.bib}

\end{document}